\documentclass[11pt,letterpaper]{article}
\usepackage{emnlp2017}
\usepackage{times}
\usepackage{latexsym}
\usepackage{algorithm,algpseudocode,algorithmicx}
\usepackage{amsmath,amssymb,amsopn,amsthm,bm,bbm}
\usepackage{booktabs}
\usepackage{graphicx}
\usepackage{siunitx}
\usepackage{tikz}
\usepackage[disable]{todonotes}
\usepackage{textcomp}
\usepackage{multirow}
\usepackage{caption}
\usepackage{subcaption}
\usepackage{float}
\usepackage{placeins}

\emnlpfinalcopy

\def\emnlppaperid{***}

\newcommand{\E}{\mathbb{E}}

\usetikzlibrary{positioning,fit,calc}
\tikzset{block/.style={draw,thick,text width=2cm,minimum height=1cm,align=center},
         line/.style={-latex}
}

\newcommand{\compresslist}{%
  \setlength{\itemsep}{1pt}%
  \setlength{\parskip}{0pt}%
  \setlength{\parsep}{0pt}%
}

\title{A Shared Task on Bandit Learning for Machine Translation}

\author{Artem Sokolov$^{\ast,\diamond}$ \and Julia Kreutzer$^\diamond$ \and Kellen Sunderland$^\ast$ \and
Pavel Danchenko$^\ast$ \\ and {\bf Witold Szymaniak$^\ast$} \and {\bf Hagen F\"urstenau$^\ast$} \and {\bf Stefan Riezler$^\diamond$}\\
{$^\ast$Amazon Development Center Germany, Berlin and $^\diamond$Heidelberg University, Germany}
}

\date{}

\begin{document}

\maketitle

\begin{abstract}
  We introduce and describe the results of a novel shared task on bandit learning for machine translation. The task was organized jointly by Amazon and Heidelberg University for the first time at the Second Conference on Machine Translation (WMT 2017). The goal of the task is to encourage research on learning machine translation from weak user feedback instead of human references or post-edits. On each of a sequence of rounds, a machine translation system is required to propose a translation for an input, and receives a real-valued estimate of the quality of the proposed translation for learning. This paper describes the shared task's learning and evaluation setup, using services hosted on Amazon Web Services (AWS), the data and evaluation metrics, and the results of various machine translation architectures and learning protocols.  

\end{abstract}

\section{Introduction}
Bandit Learning for machine translation (MT) is a framework to train and
improve MT systems by learning from weak or partial feedback: Instead of a
gold-standard human-generated translation, the learner only receives feedback
to a single proposed translation (hence the term `partial'), in form of a
translation quality judgement (a real number which can be as weak as a binary
acceptance/rejection decision). 

In the shared task, user feedback was simulated by a service hosted on Amazon
Web Services (AWS). Participants can submit translations and receive feedback
on translation quality. This is used to adapt an out-of-domain MT model,
pre-trained on mostly news texts, to a new domain (e-commerce), for the
translation direction of German (DE) to English (EN). While in our setup
feedback was simulated by evaluating a reward function on the predicted
translation against a gold standard reference, the reference translation
itself was never revealed to the learner, neither at training nor at test time. This
learning scenario has been investigated under the names of \emph{learning from
bandit feedback}\footnote{The name is inherited from a model where in each
round a gambler pulls an arm of a different slot machine (`one-armed bandit'),
with the goal of maximizing his reward relative to the maximal possible reward,
without apriori knowledge of the optimal slot machine. See
\newcite{BubeckCesaBianchi:12} for an overview.} or \emph{reinforcement
learning} (RL)\footnote{See \newcite{SuttonBarto:98} and \newcite{Szepesvari:09} for
an overview of algorithms for reinforcement learning and their relation to
bandit learning.}, and has 
important real world applications such
as online advertising \cite{ChapelleETAL:14}. In the advertising application,
the problem is to select the best advertisement for a user visiting a publisher
page. A key element is to estimate the click-through rate (CTR), i.e., the
probability that an ad will be clicked by a user so that the advertiser has to
pay. This probability is modeled by features representing user, page, and ad,
and is estimated by trading off exploration (a new ad needs to be displayed in
order to learn its click-through rate) and exploitation (displaying the ad with
the current best estimate is better in the short term) in displaying ads to
users.

\todo[color=green]{don’t understand the task of `personalized MT'. Is this different from domain adaptation?}
In analogy to the online advertising scenario, one could imagine a scenario
of personalization in machine translation where translations have to be adapted to
the user's specific purpose and domain. Similar to online advertising, where it
is unrealistic to expect more detailed feedback than a user click on a
displayed ad, the feedback in adaptive machine translation should be weaker
than a reference translation or a post-edit created by a professional
translator. Instead, the goal is to elicit binary or real-valued judgments of
translation quality from laymen users (for example, \newcite{GrahamETAL:16}
show that consistent assessments of real-valued translation quality can be provided by
crowdsourcing), or to infer feedback signals from user interactions with the
translated content on a web page (for example, by interpreting a copy-paste
action of the MT output as positive quality signal, and a correction as a
negative quality signal). The goal of this shared task is to evaluate existing
algorithms for learning MT systems from weak
feedback~\cite{sokolov15bandit,SokolovETAL:16,kreutzer2017bandit} on real-world data and compare
them to new algorithms, with a focus on performing online learning efficiently
and effectively from bandit feedback, i.e.\ the best algorithms are those that
perform fast online learning and, simultaneously, achieve high translation quality.

In the following, we present a description of the protocol and infrastructure of our online learning task, and of the data for pretraining, online training, and evaluation (Section \ref{sec:task}). We introduce the online and batch evaluation metrics used in the shared task (Section \ref{sec:eval}), and describe static baseline systems (Section \ref{sec:baselines}) and submitted online learning systems (Section \ref{sec:systems}). We present and discuss the results of the task (Section \ref{sec:results}-\ref{sec:conclusion}), showing that NMT systems with batch domain adaptation provide very good baselines, however, online learning based on SMT or NMT can catch up over time by adapting to the provided feedback.

\section{Task Description}
\label{sec:task}
Our shared task setup follows an online learning protocol, where on each
iteration, the learner receives a source sentence, proposes a translation, and
is rewarded in form of a task sentence-level metric evaluation of the proposed
translation with respect to a hidden reference. The learner does not know what
the correct translation (reference) looks like, nor what would have happened if
it had proposed a different translation. Thus, we implemented two constraints
to guarantee this scenario of online learning from weak feedback. First,
sentences had to be translated one by one, i.e. the next source sentence could
only be received after the translation to the previous sentence was sent off.
Second, feedback could be obtained only for a single translation of any given source
sentence.  

In our shared task, the participant systems interact online with an AWS-hosted service
as shown in Algorithm~\ref{alg:bandit}. 
The service provides a source sentence to the learner (line~3), and provides
feedback (line~5) to the translation predicted by the learner (line~4). The
learner updates its parameters using the feedback (line~6) and continues to the
next example.  We did not impose any restriction on how the learner could use
the feedback to improve future translations.
 
\begin{algorithm}[t]
\caption{WMT Online Bandit Learning}
\label{alg:bandit}
\begin{algorithmic}[1]
\algnotext{EndFor}
\State Input: MT model
\For{$k=0,\ldots,K$}
\State Request source sentence $s_k$ from service
\State Propose a translation $t_k$
\State Obtain feedback $\Delta(t_k)$ from service
\State Improve MT model with $\langle s_k, t_k, \Delta(t_k) \rangle$
\EndFor
\end{algorithmic}
\end{algorithm}

\begin{table*}[t]
\resizebox{\textwidth}{!}{%
\begin{tabular}{ll|cc}
\toprule
\bf source & \bf reference (PE) & \bf PE direction & \bf PE modification\\
\midrule
schwarz gr.xxl / xxxl & black , size xxl / xxxl & DE-EN & fixed errors in source, expanded abbreviation\\
, 147 cm & 147 cm & DE-EN & fixed errors in source \\
f\"{u}r starke , gl\"{a}nzende n\"{a}gel & great for strengthen your nails and enhance shine & EN-DE & poor quality source (EN) used as reference\\
seemless verarbeitung & seamless processing & DE-EN & source typo corrected in reference\\
brenndauer : mindestens 40 stunden & 40 hour minimum burn time & DE-EN & translation rewritten for readability\\
maschinenwaschbar bei 30 \textdegree~c & machine washable at 30 degrees . &DE-EN& literal expansion of the degree symbol\\
32 unzen volumen & 32-ounce capacity & DE-EN & language-specific typography\\
material : 1050 denier nylon . & material : 1050d nylon . & EN-DE & expanded source (EN) abbreviation used as reference\\ 
f\"{u}r e-gitarre entworfen & designed for electric guitar & DE-EN & abbreviation expanded\\
\bottomrule
\end{tabular}
}
\caption{Examples for non-literal PEs in the e-commerce data: The first two columns show examples\footnotemark\ of source sentences and PEs used as reference translations in the shared task. The last two columns show the direction of translation post-editing, and a description of the modifications applied by the editors.}
\label{tab:examples}
\end{table*}

\paragraph{Infrastructure.}

We provided three AWS-hosted environments, that correspond to the three phases of the shared task:
\begin{enumerate}\compresslist
\item Mock service, to test the client API (optional): hosted a tiny in-domain dataset (48 sentences).
\item Development service to tune algorithms and hyperparameters (optional): ran on a larger in-domain dataset (40,000 sentences). Several passes were allowed and two evaluation metrics were communicated to the participants via the leaderboard.
\item Training service (mandatory): served sources from a large in-domain dataset (1,297,974 sentences). Participants had to consume a fixed number of samples during the allocated online learning period to be eligible for final evaluation. 
\end{enumerate} 

We built the shared task around the following AWS services:
\begin{itemize}\compresslist
    \item API Gateway (authentication, rate limiting, client API SDK);
    \item Lambda (computation);
    \item DynamoDB (data storage);
    \item CloudWatch (logging and monitoring).
\end{itemize}
In more detail, service
endpoints were implemented using API Gateway, that gave us access, on a
participant level, to throttle requests rates, manage accounts, etc. API
Gateway enabled easy management of our public-facing endpoints and
environments, and provided integrated metrics and notifications, which we
monitored closely during the shared task. Data storage was implemented using
DynamoDB -- a NoSQL storage database which allows dynamic scaling of our
back-end to match the varied requirements of the different shared task phases.
The state management (e.g., forbidding multiple requests), source sentence
serving, feedback calculation, keeping track of participant's progress and
result processing was implemented using Lambda -- a serverless compute
architecture that dispenses with setting up and monitoring a dedicated server
infrastructure.  CloudWatch service was used to analyze logs in order to trace down 
errors, general monitoring and sending alarms to the shared task API maintainers.
In addition to the service development, we also developed a small SDK
consisting of code samples and helper libraries in Python and Java to help
participants in developing their clients, as well as a leaderboard that showed
the results during the development phase.

\paragraph{Data.} 

\footnotetext{Examples selected by Khanh Ngyuen.}
For training initial or \emph{seed} MT systems (the input to Algorithm~\ref{alg:bandit}), out-of-domain parallel data was restricted to DE-EN
parts of Europarl~v7, NewsCommentary v12, CommonCrawl and Rapid data from the WMT 2017
News Translation (constrained)
task\footnote{\mbox{\url{statmt.org/wmt17/translation-task.html}}}. Furthermore,
monolingual EN data from the constrained task was allowed. Tuning of the
out-of-domain systems had to be done on the \texttt{newstest2016-deen}
set.

The in-domain parallel data for online learning was taken from the e-commerce
domain: The corpus was provided by Amazon and had been sampled from a large
real-world collection of post-edited (PE'ed) translations of actual product
descriptions.  Since post-editors were following certain rules aimed at
improving customer experience on the Amazon retail website (improving readability,
correction of typos, rewriting of uncommon abbreviations, removing irrelevant
information, etc.), naturally the resulting PEs were not
always literal, sometimes adding or deleting a considerable number of tokens
and resulting in low feedback BLEU scores for submitted literal translations
(see Table~\ref{tab:examples} for examples).  Consequently, the participants
had to solve two difficult problems -- domain adaptation and learning from
bandit feedback. In addition, to simulate the level of noise normally
encountered in real-world MT applications, and to test noise-robustness of the
bandit learning algorithms, approximately half of the parallel in-domain data
was sourced from the EN-DE post-editing direction and reversed.

All data was preprocessed with \texttt{Moses} scripts (removing non-printing
characters, replacing and normalizing unicode punctuation, lowercasing,
pretokenizing and tokenizing). No DE-side compound splitting was used, permitting
custom participant decisions.  Since the learning data came from a substantially
different domain than the out-of-domain parallel texts, it had a large
number of out-of-vocabulary (OOV) terms, aggravated by the high frequency of
long product numbers and unique vendor names. To reduce the OOV rate we
additionally filtered out all parallel sentences where the source contained
more than one numeral (with a whitespace in between) and normalized floating point
delimiters in both languages to a period.  The resulting
average OOV token rate with respect to the out-of-domain parallel training data (assuming
the above preprocessing) is $\simeq 2$\% for EN and $\simeq 6$\% for DE data
side. Statistics on the length distribution of in-domain and out-of-domain data is given in Table~\ref{tab:stats}.

For all services, the sequence of provided source sentences was the same for all participants, with no data intersection between the services beyond
natural duplicates: About 11\% of data were duplicates on
both (DE and EN) sides, where about 4\% of DE sentences had more than one
different EN side.

\paragraph{Feedback.} Simulation of real-valued user feedback was done by calculating
the smoothed sentence-level BLEU-score~\cite{lin04auto} (with additive $n$-gram count smoothing with offset 0.01, applied only if the $n$-gram count was zero) with respect to one human reference
(preprocessed as described above).
\todo[color=green]{Do you want to say anything about us considering other metrics than BLEU, but deciding against switching between dev and test phase, since the task seemed hard enough already?}

\begin{table}[t]
\centering
\begin{tabular}{lcc}
\toprule
\bf \# tokens & \bf out-of-domain & \bf in-domain \\
\midrule
mean     & 23.0$\pm$14.1 & 6.6$\pm$4.8\\
median   & 25            & 8\\
max      & 150           & 25\\
\midrule
\midrule
\bf \# lines      & 5.5M           & 1.3M\\
\bottomrule
\end{tabular}
\caption{Data statistics for source side of in-domain and out-of-domain parallel data.}
\label{tab:stats}
\end{table}

\section{Evaluation Metrics}
\label{sec:eval}
In our shared task, participants were allowed to use their favorite MT systems
as starting points to integrate online bandit learning methods. This leads to
the difficulty of separating the contributions of the underlying MT
architecture and the online bandit learning method. We attempted to tackle this
problem by using different evaluation metrics that focus on these respective
aspects:

\begin{enumerate}
 \item \textbf{Online cumulative reward}: This metric measures the
cumulative sum $C = \sum_{k=1}^K \Delta(t_k)$ of the per-sentence BLEU score
$\Delta$ against the number of iterations. This metric has been used in
reinforcement learning competitions~\cite{dimitrakakis2014reinforcement}. For
systems with the same design, this metric favors those that do a good job at
balancing exploration and exploitation to achieve high scores over the full
data sequence. Unlike in these competitions, where environments (i.e., action
spaces and context features) were fixed, in our task the environment is
heterogeneous due to the use of different underlying MT architectures. Thus,
systems that start out with a well-performing pretrained out-of-domain model
have an advantage over systems that might improve more over worse starting
points. Furthermore, even systems that do not perform online learning at all
can achieve high cumulative rewards.
 
\item \textbf{Online regret:} In order to overcome the problems of the
cumulative reward metric, we can use a metric from bandit learning that
measures the regret $R=\frac{1}{K}\sum_{k=1}^K \big(\Delta(t^\ast_k) -
\Delta(t_k)\big)$ that is suffered by the system when predicting translation
$t_k$ instead of the optimal translation $t^\ast_k$ produced by an oracle system.
Plotting a running average
of regret against the number of iterations allows separating the gains due to
the MT architecture from the gains due to the learning algorithm: Systems that
do learn will decrease regret, systems that do not learn will not. In our task,
we use as oracle system 
a model that is trained on in-domain data. 

 \item \label{enum:bleu} \textbf{Relative reward}: A further way to
separate out the learning ability of systems from the contribution of the
underlying MT architecture is to apply the standard corpus-BLEU score and/or an average of the per-sentence BLEU score $\Delta$ on a held-out set at regular intervals during training. Plotting these scores against the number of iterations, or alternatively, subtracting the performance of the starting point at each evaluation, allows to discern systems that adapt to a
new domain from systems that are good from the beginning and can
achieve high cumulative rewards without learning.  We performed this evaluation
by embedding a small (relative to the whole sequence) fixed held-out set in the
beginning (showing the performance of the initial out-of-domain model),
and again at regular intervals including the very end of the learning
sequence. In total, there were 4 insertions of 700 sentences in the development
data and 12 insertions of 4,000 sentences in the final training phase, which
constitutes $\simeq$2\% and $\simeq$0.3\% of the respective learning sequence lengths. Note
that this metric measures the systems' performance while they were still
exploring and learning, but the relative size of the embedded held-out set is small
enough to consider the models static during such periodic evaluations.
\end{enumerate}

\section{Baselines}
\label{sec:baselines}
As baseline systems, we used SMT and NMT models that were trained on out-of-domain data, but did not perform online learning on in-domain data. We further present oracle systems that were trained in batch on in-domain data.

\subsection{Static SMT baselines.} 

\paragraph{SMT-static.}
We based our SMT submissions on the SCFG decoder
\texttt{cdec}~\cite{DyerETAL:10} with on-the-fly grammar extraction with suffix
arrays~\cite{lopez07}. Training was done in batch on the parallel out-of-domain data; tuning was done on \texttt{newstest2016-deen}. During
the development phase we evaluated MERT (on 14 default dense features) and MIRA
(on additional lexicalized sparse features: rule-id features, rule source and
target bigram features, and rule shape features), and found no significant
difference in results. We chose MERT with dense features as the seed system for the training phase
for its speed and smaller memory footprint.

\subsection{Static NMT baselines.}

\paragraph{WMT16-static.}
First of all, we are interested in how well the currently best (third-party) \todo{added 3party}
model on the 
news domain would perform on the e-commerce domain. Therefore, the Nematus
\cite{sennrich-EtAl:2017:EACLDemo} model that won the News
Translation Shared Task at WMT 2016
\cite{bojar-EtAl:2016:WMT1}\footnote{From
\url{data.statmt.org/rsennrich/wmt16_systems/de-en/}} was used to
translate the data from this shared task. It is an attentional, bi-directional,
singe-layered encoder-decoder model on sub-word units (BPE with 89,500 merge
operations) with word embeddings of dimensionality 500, GRUs of size 1024,
pervasive dropout and r2l reranking (details in \cite{sennrich-haddow-birch:2016:WMT}). Final predictions are made with an
ensemble formed of the four last training checkpoints and beam search with width 12. It was trained on a different corpus than allowed for this shared task -- the WMT 2016 news training data (Europarl v7, News Commentary v11, CommonCrawl) and additional synthetic parallel data generated by translating the monolingual news crawl corpus with a EN-DE NMT model.

\paragraph{BNMT-static.}
The UNK replacement strategy of \citet{jean-EtAl:2015:WMT} and \citet{luong-EtAl:2015:ACL-IJCNLP} is expected to work reasonable well for tokens
that occur in the training data and those that are copied from source to target.
However, the NMT model does not learn anything about these words as such in
contrast to BPE models \cite{sennrich-haddow-birch:2016:P16-12} where the decomposition by byte pair encoding (BPE) allows for a representation
within the vocabulary. We generate a BNMT system using a BPE vocabulary from 30k merge operations on all tokens and all single characters of the training data, including the UNK token.  If unknown characters occur, they are copied from source to target. 

\subsection{Oracle SMT and NMT systems}
To simulate full-information systems (oracles) for regret
calculation, we trained an SMT and an NMT system with the same architectures,
on the in-domain data that other learning systems accessed only through the
numerical feedback. The SMT oracle system was trained on combined in-domain and
out-of-domain data, while the NMT oracle system 
continued training from the converged out-of-domain system on the in-domain data 
with the same BPE vocabulary. 

\section{Submitted Systems}
\label{sec:systems}
\subsection{Online bandit learners based on SMT.}
Online bandit learners based on SMT were following the existing approaches to adapting an SMT model from weak user feedback~\cite{sokolov16bandit,SokolovETAL:16} by
stochastically optimizing expected loss (EL) for a log-linear model. Furthermore, we present a model that implements stochastic zeroth-order (SZO) optimization for online bandit learning.
Cube pruning limit (up to 600), learning rate adaptation schedules (constant vs.\ Adadelta~\cite{zeiler12adadelta} or
Adam~\cite{kingma2014adam}), as well as the initial learning rates (for Adam),
were tuned during the development phase. The best configurations were selected
for the training phase. The running average of rewards as an additive control
variate (CV)\footnote{Called a \emph{baseline} in RL literature; here we use a
term from statistics not to confuse it with baseline MT models.} was found
helpful for stochastic policy gradient updates~\cite{Williams:92} for all
online learning systems.

\paragraph{SMT-EL-CV-ADADELTA.} We used the EL minimization approach of~\newcite{SokolovETAL:16}, adding Adadelta's learning rate scheduling, and a control variate
(effectively, replacing the received feedback $\Delta(t_k)$ with
$\Delta(t_k)-\frac{1}{k}\sum_{k'=1}^k\Delta(t_{k'})$). Sampling and computation
of expectations on the hypergraph used the Inside-Outside
algorithm~\cite{li2009first}.

\paragraph{SMT-EL-CV-ADAM.}
This system uses the same approach as above except for using Adam to adapt the learning rates, with tuning of the initial learning rate on the development service.

\paragraph{SMT-SZO-CV-ADAM.} 
As a novel contribution, we adapted the \emph{two-point} stochastic
zeroth-order approach by~\cite{sokolov15bandit} that required two quality
evaluation per iteration to a \emph{one-point} feedback scenario. In a
nutshell, on each step of the SZO algorithm, the model parameters $w$ are
perturbed with an additive standard Gaussian noise $\epsilon$, and the Viterbi
translation is sent to the service. Such algorithm can be shown to maximize the
smoothed version of the task reward: $\E_{\epsilon\sim N(0,1)}[\Delta(\hat
y(w+\epsilon))]$~\cite{FlaxmanETAL:05}. The advantages of such a black-box optimization method over model-based (e.g.\ EL) optimization, that requires
sampling of complete structures from the model distribution, are simpler
sampling of standard Gaussians, and matching of the inference criterion to the
learning objective (MAP inference for both), unlike the EL optimization of
\emph{expected} reward that is still evaluated at test time using MAP inference. For SZO models we found that the Adam scheduling consistently outperforms Adadelta.

\subsection{Online bandit learners based on NMT.}

\citet{kreutzer2017bandit} recently presented an algorithm for online expected
loss minimization to adapt NMT models to unknown domains with bandit
feedback. Exploration (i.e.\ sampling from the model) and exploitation (i.e.\ presenting the highest scored translation) are controlled by the softmax
distribution in the last layer of the network. Ideally, the model would converge
towards a peaked distribution. In our online learning scenario this is not
guaranteed, but we would like the model to gradually stop exploring, in order
to still achieve high cumulative per-sentence reward. To achieve such a
behavior, the temperature of the softmax over the outputs of the last layer of the network is annealed \cite{Rose:98}. More specifically, let $o$ be the scores of the output projection layer of the decoder, then $p_{\theta}(\tilde y_t = w_i |\mathbf{x}, \mathbf{\hat y}_{< t}) = \frac{\exp(o_{w_i}/T)}{\sum_{v=1}^{V}{\exp(o_{w_v}/T)}}$ is the distribution that defines the probability of each word $w_i$ of the target vocabulary $V$ to be sampled in timestep $t$. The annealing schedule for this temperature $T$ is defined as $T_k =
0.99^{\max(k-k_{\text{START}}, 0)}$, i.e. decreases from iteration
$k_{\text{START}}$ on. The same decay is applied to the learning rate, such
that $\gamma_k = \gamma_{k-1}\cdot T_k$. This schedule was proven successful
during tuning with the leaderboard. 


\paragraph{WNMT-EL.}
Using the implementation of \citet{kreutzer2017bandit}, we built a word-based NMT system with NeuralMonkey \cite{libovicky2016cuni, bojar2016ufal} and trained it with the EL algorithm. The vocabulary is limited to the 30k most frequent words in the out-of-domain training corpus. The architecture is similar to WMT16-static with GRU size 1024, embedding size 500. It was pretrained on the out-of-domain data with the standard maximum likelihood objective, Adam ($\alpha = \num{1e-4}$, $\beta_1 = 0.9$, $\beta_2 = 0.999$) and dropout \cite{srivastava2014dropout} with probability 0.2. 
Bandit learning starts from this pretrained model and continues with stochastic gradient descent (initial learning rate $\gamma_0 = \num{1e-5}$, annealing starts at $k_{START} = 700,000$, dropout with probability 0.5, gradient norm clipping when the norm exceeds 1.0 \cite{pascanu2013difficulty}), where the model was updated as soon as a feedback is received. As described above, UNK replacement was applied to the output on the basis of an IBM2 lexical translation model built with \texttt{fast\_align}~\cite{dyer13simple} on out-of-domain training data. If the aligned source word for a generated UNK token is not in the dictionary of the lexical translation model, the UNK token was simply replaced by the source word.

\paragraph{BNMT-EL.}
The pretrained BPE model is further trained on the bandit task data with the EL algorithm, as described for BL1, with the only difference of using Adam ($\alpha = \num{1e-5}$, $\beta_1 = 0.9$, $\beta_2 = 0.999$) instead of SGD. Again, annealing started at $k_{START} = 700,000$.

\paragraph{BNMT-EL-CV.}
BNMT-EL-CV is trained in the same manner as BNMT-EL with the addition of the same control variate technique
(running average of rewards) that has been previously found to improve both variance and generalization
for NMT bandit training~\cite{kreutzer2017bandit}.

\subsection{Domain adaptation and reinforcement learning based on NMT (University of Maryland).}

\paragraph{UMD-domain-adaptation.} The UMD team's systems were based on an attention-based encoder-decoder translation model. The models use the BPE technique for subword encoding, which helps addressing the rare word problem and enlarges vocabulary. A further addition is the domain adaptation approach of \citet{axelrod11} to select training data after receiving in-domain
source-side data and selecting the most similar out-of-domain data from the WMT~2016
training set for re-training.

\paragraph{UMD-reinforce.} Another type of models submitted by UMD uses reinforcement learning techniques to learn from feedback and improve the update of the translation model to optimize the reward, based on~\newcite{bahdanau16} and~\newcite{RanzatoETAL:16}.

\subsection{Domain adaptation and bandit learning based on SMT (LIMSI).}

\paragraph{LIMSI.}
The team from LIMSI tried to adapt a seed \texttt{Moses} system trained on out-domain
data to a new, unknown domain relying on two components, each of which
addresses one of the challenges raised by the shared task: i) estimate
the parameters of a MT system without knowing the reference
translation and in a `one-shot' way (each source sentence can only be
translated once); ii) discover the specificities of the target domain
`on-the-fly' as no information about it is available.
First, a linear regression model was used to exploit weak and
partial feedback the system received by learning to predict the reward
a translation hypothesis will get. This model can then be used to
score hypotheses of the search space and translate source sentences
while taking into account the specificities of the in-domain
data. Second, three variants of the UCB1~\cite{auer} algorithm (vanilla UCB1, a UCB1-sampling variant encouraging more exploration, and a UCB1 with selecting only the examples not used to train the regression model) chose which of the `adapted' or `seed' systems should be used to translate a given source
sentence in order to maximize the cumulative reward~\cite{wisniewski17}.

\section{Results}
\label{sec:results}
\begin{table}[t]
\resizebox{\columnwidth}{!}{
\begin{tabular}{cl|c}
\toprule
& \multirow{2}{*}{\bf model} & \bf cumulative\\ 
&& \bf reward \\
\midrule
& `translate' by copying source & ~~64,481.8 \\
\midrule
\midrule
\multirow{5}{*}{\rotatebox{90}{SMT}} & SMT-oracle & 499,578.0 \\
& SMT-static & 229,621.7\\
\cmidrule(r){2-3}
&SMT-EL-CV-ADADELTA                & 214,398.8\\
&SMT-EL-CV-ADAM     & 225,535.3\\
&SMT-SZO-CV-ADAM & 208,464.7\\
\midrule
\midrule
\multirow{6}{*}{\rotatebox{90}{NMT}}& BNMT-oracle & 780,580.4\\
& BNMT-static & 222,066.0\\
& WMT16-static & 139,668.1\\
\cmidrule(r){2-3}
& BNMT-EL-CV & 212,703.2\\
& BNMT-EL & 237,663.0\\
& WNMT-EL & 115,098.0\\
\midrule
\midrule
& UMD-domain-adaptation & 248,333.2\\
\bottomrule
\end{tabular}
}
\caption{Cumulative rewards over the full training sequence. Only completely finished submission are shown.}
\label{tab:results}
\end{table}

Table \ref{tab:results} shows the evaluation results under the cumulative rewards metric. Of the non-oracle systems, good results are obtained by static SMT and BNMT system, while the best performance is obtained by the UMD-domain adaptation system which is also basically a static system. This result is followed closely by the online bandit learner BNMT-EL which is based on an NMT baseline and optimizes the EL objective. It outperforms the BNMT-static baseline. Cumulative rewards could not be computed for all submitted systems since some training runs could not be fully finished.

\begin{figure*}[t]
\vspace{-4em}
\includegraphics[width=\textwidth]{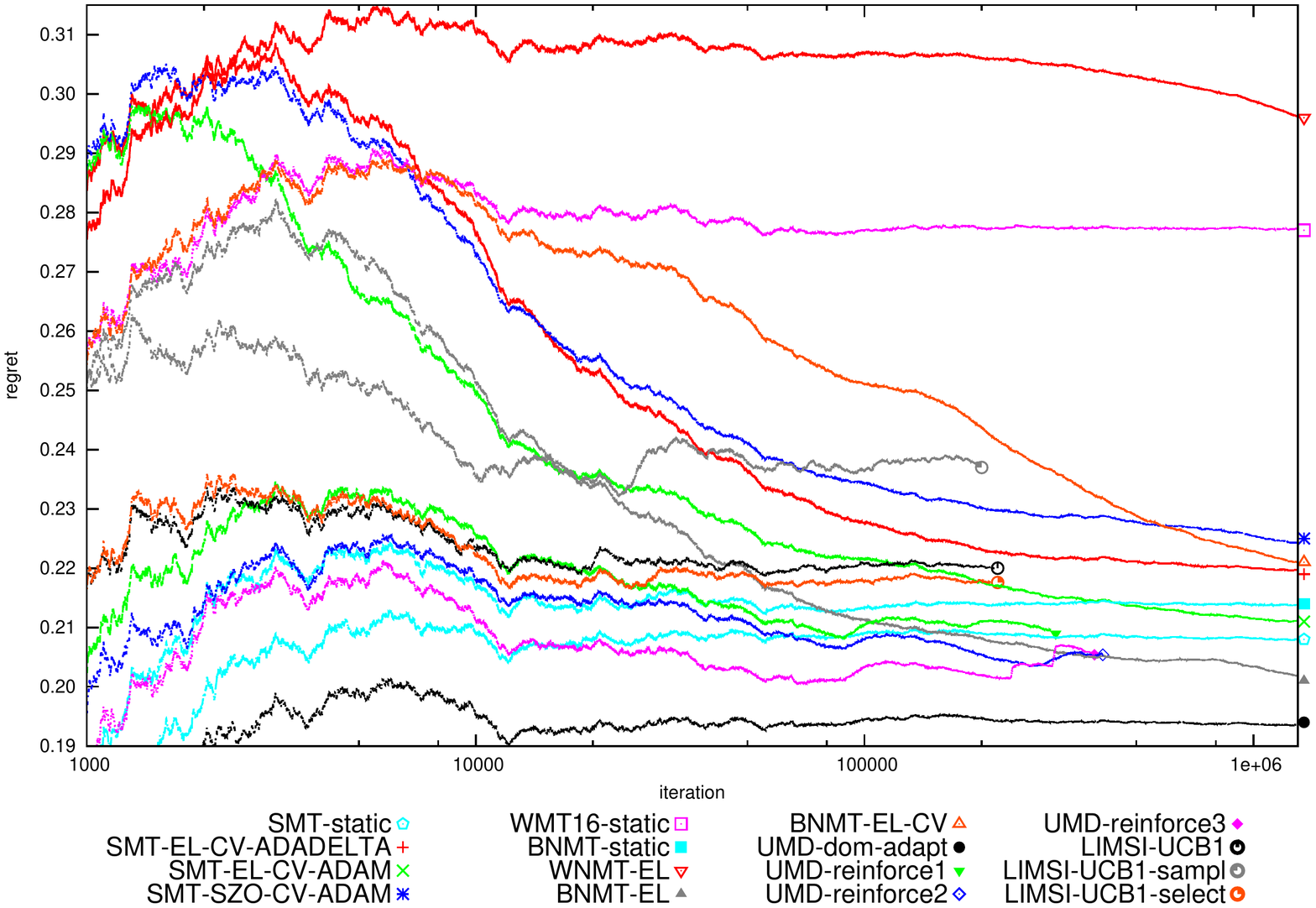}
\vspace{-3em}
\caption{Evolution of regret plotted against log-scaled number of iterations
during training. The steeper is the decrease of a curve, the better learning capability has the corresponding algorithm. \label{fig:regret}}
\vspace{-1em}
\includegraphics[width=\textwidth]{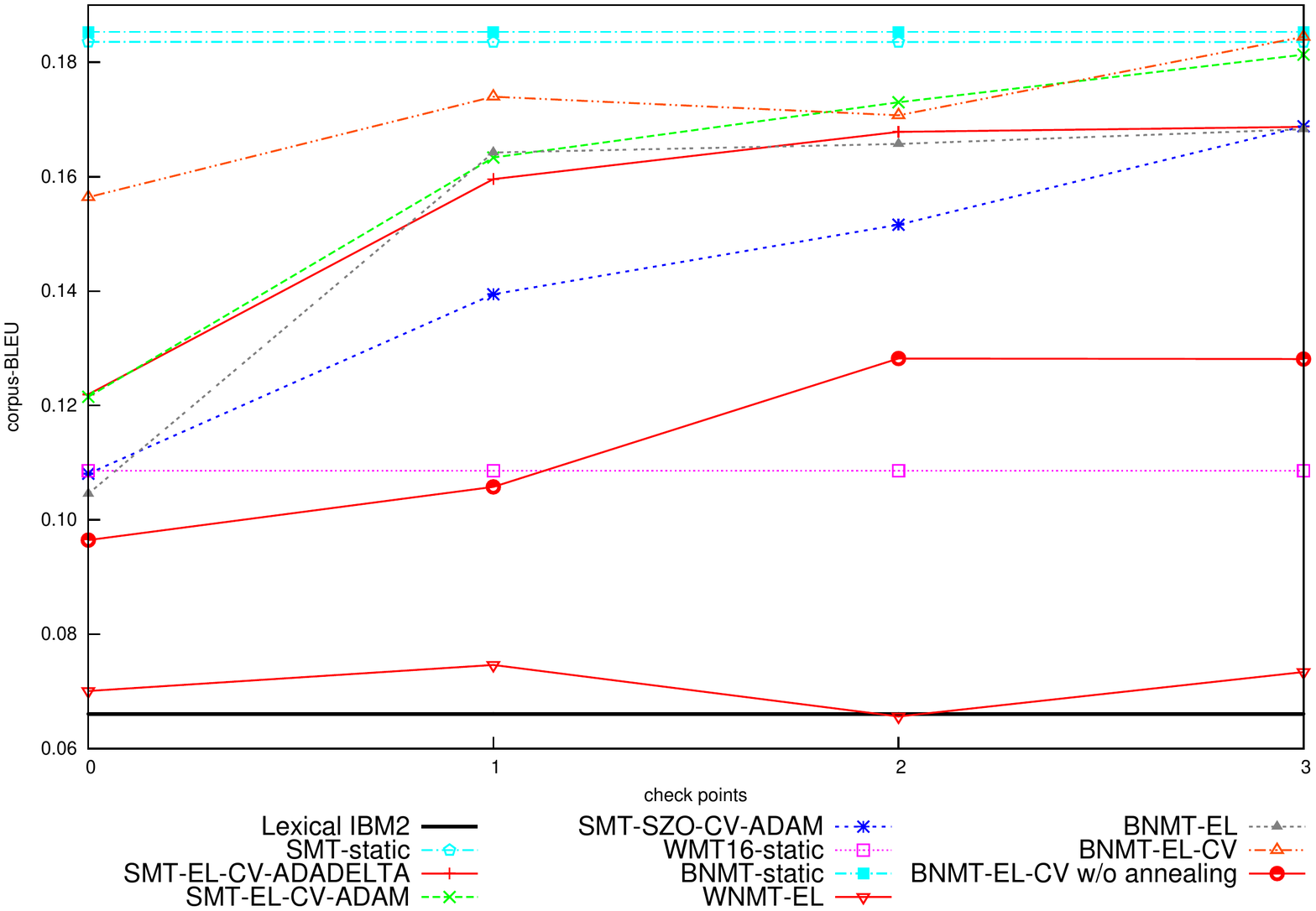}
\vspace{-3em}
\caption{Evolution of corpus BLEU scores during development for
configuration selected for the training phase of the competition. Each check
point is comprised of the same 700 sentences spaced at a regular intervals of
12,400 sentences starting from the beginning of the development sequence.\label{fig:dev}}
\end{figure*}

The evolution of the online regret plotted against the log-scaled
number of iterations during training is shown in Figure~\ref{fig:regret}. Most
of the learning happens in the first 100,000 iterations, however, online
learning systems optimizing structured EL objectives or based on reinforcement learning eventually converge to the same result: BNMT-EL or UMD-reinforce2 get close to the regret of the static UMD-domain adaptation. Systems that
optimize the EL objective do not start from strong out-of-domain
systems with domain-adaptation, however, due to a steeper learning
curve they arrive at similar results.

\begin{figure*}[t]
\vspace{-3em}
\begin{subfigure}[t]{\textwidth}
\includegraphics[width=\textwidth]{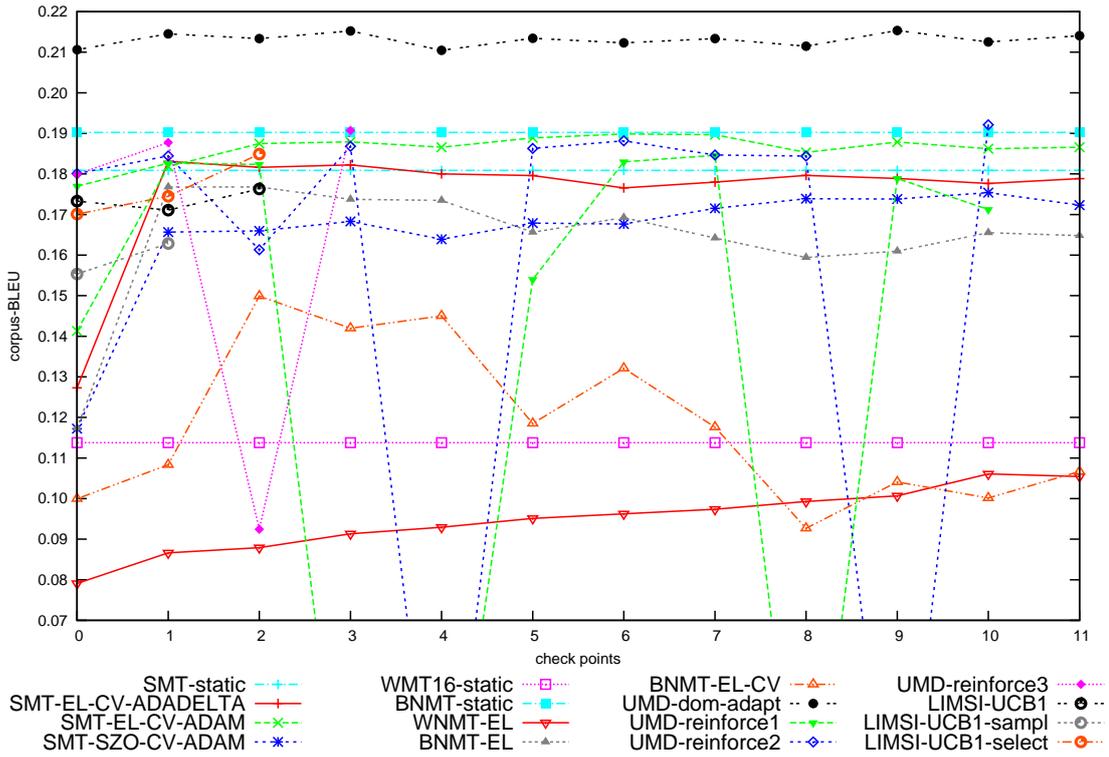}
\vspace{-3em}
\caption{corpus-BLEU}
\label{fig:train_corpus_bleu}
\end{subfigure}
\\
\begin{subfigure}[t]{\textwidth}
\vspace{-3em}
\includegraphics[width=\textwidth]{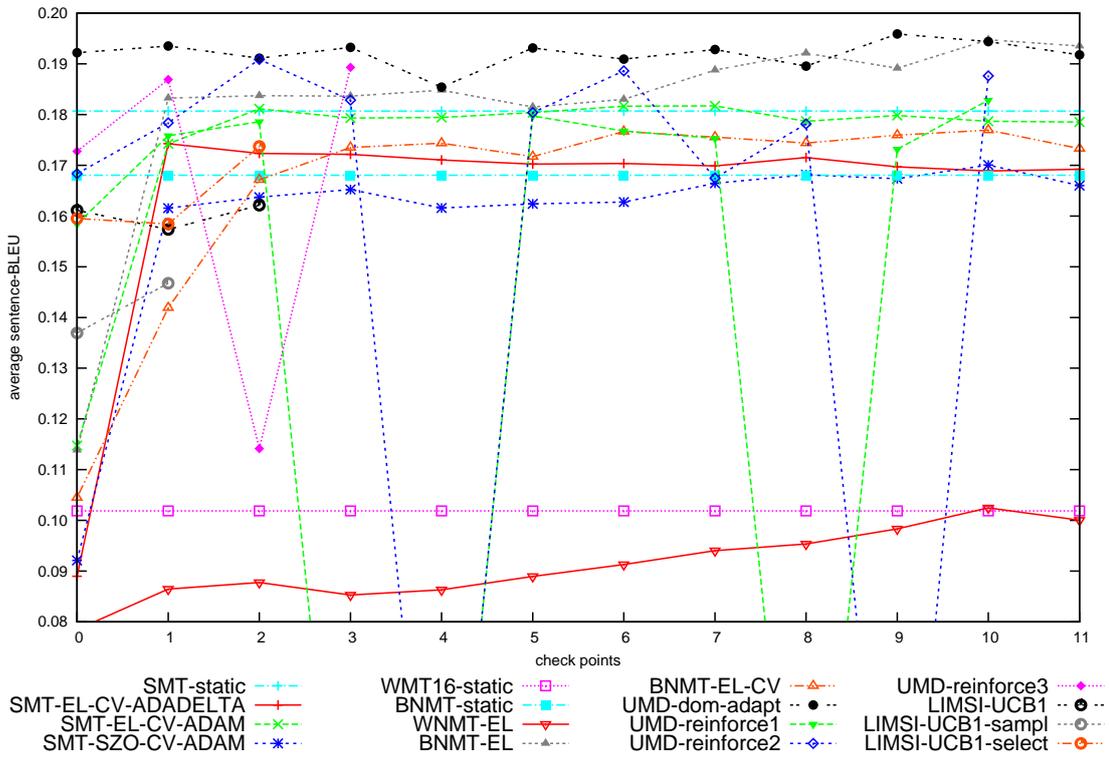}
\vspace{-3em}
\caption{sentence-BLEU}
\label{fig:train_sent_bleu}
\end{subfigure}
\caption{The evolution of corpus- and sentence-BLEU scores during training for all participant and baselines.
Each check point is comprised of the same 4,000 sentences spaced at a regular intervals of
113,634 sentences starting from the beginning of the training sequence.
}
\label{fig:train_bleu}
\end{figure*}

Figures \ref{fig:dev}, \ref{fig:train_corpus_bleu} and \ref{fig:train_sent_bleu}
show the evolution of corpus-
and sentence-BLEU on the heldout set that \FloatBarrier \noindent has been embedded in the development and the training sequences. \todo[color=green]{UMD-domain outperms all, but it wiggles so not static?} While under corpus-BLEU, static systems always outperform online
learners on the held-out embedded set, online learning systems such as BNMT-EL
can catch up under corpus-BLEU during development, and under a
sentence-BLEU evaluation during training.
The curves for corpus- and average
sentence-BLEU (Figures~\ref{fig:train_corpus_bleu}
and~\ref{fig:train_sent_bleu}) show a different dynamics, with the corpus-BLEU
sometimes decreasing whereas the sentence-BLEU curve continues to increase. 
However, if the focus is
online learning, the online task loss is per-sentence BLEU and so should be the
evaluation metric.

\section{Conclusion}
\label{sec:conclusion}
We presented the learning setup and infrastructure, data and evaluation metrics, and descriptions of baselines and submitted systems for a novel shared task on bandit learning for machine translation. The task implicitly involved domain adaptation from the news domain to e-commerce data (with the additional difficulty of non-literal post-editions as references), and online learning from simulated per-sentence feedback on translation quality (creating a mismatch between the per-sentence task loss and the corpus-based evaluation metric standardly used in evaluating batch-trained machine translation systems). Despite these challenges, we found promising results for both linear and non-linear online learners  that could outperform their static SMT and NMT baselines, respectively. A desideratum for a future installment of this shared task is the option to perform \emph{offline} learning from bandit feedback \cite{LawrenceETAL:17}, thus allowing a more lightweight infrastructure, and opening the task to (mini)batch learning techniques that are more standard in the field of machine translation.

\section*{Acknowledgments}

This research was supported in part by the German research foundation (DFG), and in part by a research cooperation grant with the Amazon Development Center Germany. We would like to thank Amazon for supplying data and engineering expertise, and for covering the running costs.

\bibliography{bib}
\bibliographystyle{emnlp_natbib}


\end{document}